\documentclass[sigconf]{acmart}
\AtBeginDocument{%
  }

\copyrightyear{2026}
\acmYear{2026}
\setcopyright{cc}
\setcctype{by}
\acmConference[ICMR '26]{International Conference on Multimedia Retrieval}{June 16--19, 2026}{Amsterdam, Netherlands}
\acmBooktitle{International Conference on Multimedia Retrieval (ICMR '26), June 16--19, 2026, Amsterdam, Netherlands}
\acmDOI{10.1145/3805622.3810595}
\acmISBN{979-8-4007-2617-0/2026/06}
\usepackage{multirow}
\usepackage{float}
\usepackage{algorithm}
\usepackage{algpseudocode}
\usepackage{booktabs}

\begin{document}

%%
%% The "title" command has an optional parameter,
%% allowing the author to define a "short title" to be used in page headers.
\title{From Diffusion to Rectified Flow: Rethinking Text-Based Segmentation}

%%
%% The "author" command and its associated commands are used to define
%% the authors and their affiliations.
%% Of note is the shared affiliation of the first two authors, and the
%% "authornote" and "authornotemark" commands
%% used to denote shared contribution to the research.
% \author{Ben Trovato}
% \authornote{Both authors contributed equally to this research.}
% \email{trovato@corporation.com}
% \orcid{1234-5678-9012}
% \author{G.K.M. Tobin}
% \authornotemark[1]
% \email{webmaster@marysville-ohio.com}
% \affiliation{%
%   \institution{Institute for Clarity in Documentation}
%   \city{Dublin}
%   \state{Ohio}
%   \country{USA}
% }

\author{Zishen Qu}
\affiliation{%
  \institution{Zhejiang University}
  \city{Beijing}
  \country{China}}
\email{wangyang.333@bytedance.com}

\author{Xuesong Li}
\affiliation{%
  \institution{bytedance}
  \city{Beijing}
  \country{China}}
\email{lixuesong.darkn@bytedance.com}

\author{Haijian Gu}
\affiliation{%
  \institution{Southeast University}
  \city{Nanjing}
  \country{China}}
\email{220232244@seu.edu.cn}

\author{Quan Meng}
\affiliation{%
  \institution{bytedance}
  \city{Shanghai}
  \country{China}}
\email{mengquan@bytedance.com}

\author{Tianrui Niu}
\affiliation{%
  \institution{bytedance}
  \city{Beijing}
  \country{China}}
\email{niutianrui@bytedance.com}

\author{Xin Yang}
\affiliation{%
  \institution{bytedance}
  \city{Beijing}
  \country{China}}
\email{yangxin.yangxin@bytedance.com}

\author{Ruidong Pan}
\affiliation{%
  \institution{bytedance}
  \city{Shanghai}
  \country{China}}
\email{prd@bytedance.com}

\author{Hongwei Kang\authornotemark}
\affiliation{%
  \institution{bytedance}
  \city{Beijing}
  \country{China}}
\email{hongwei.khw@bytedance.com}

\authornote{Corresponding author: Hongwei Kang (hongwei.khw@bytedance.com)}
% \cortext[cor]{Corresponding author.} 
%%
%% By default, the full list of authors will be used in the page
%% headers. Often, this list is too long, and will overlap
%% other information printed in the page headers. This command allows
%% the author to define a more concise list
%% of authors' names for this purpose.
% \renewcommand{\shortauthors}{Trovato et al.}

%%
%% The abstract is a short summary of the work to be presented in the
%% article.
\begin{abstract}
  A clear and well-documented \LaTeX\ document is presented as an
  article formatted for publication by ACM in a conference proceedings
  or journal publication. Based on the ``acmart'' document class, this
  article presents and explains many of the common variations, as well
  as many of the formatting elements an author may use in the
  preparation of the documentation of their work.
\end{abstract}

%%
%% The code below is generated by the tool at http://dl.acm.org/ccs.cfm.
%% Please copy and paste the code instead of the example below.
%%


%%
%% Keywords. The author(s) should pick words that accurately describe
%% the work being presented. Separate the keywords with commas.

% \received{20 February 2007}
% \received[revised]{12 March 2009}
% \received[accepted]{5 June 2009}

%%
%% This command processes the author and affiliation and title
%% information and builds the first part of the formatted document.
\begin{abstract}
Text-based image segmentation aims to delineate object boundaries within an image from text prompts, offering higher flexibility and broader application scope compared to traditional fixed-category segmentation tasks. 
Recent studies have shown that diffusion models (e.g., Stable Diffusion) can provide rich multimodal semantic features, leading to studies of using diffusion models as feature extractors for segmentation tasks. Such methods, however, inherit the generative natures of diffusion models that are harmful to discriminative segmentation tasks. In response, we propose RLFSeg, a novel framework that leverages Rectified Flow to learn direct mapping from the image to the segmentation mask within the latent space. The model is thus freed from the noise-denoise process and the need to optimize the time step of diffusion models, resulting in substantially better performance than previous diffusion-based methods, especially on zero-shot scenarios. By introducing label refinement and an Adaptive One-Step Sampling strategy, the model achieves higher accuracy even on a single inference step. The framework redirects a pretrained generative model to the discriminative segmentation task with zero modification to model structure, thus reveals promising application potential and significant research value.
\end{abstract}

\maketitle
\section{INTRODUCTION}
Text-based image segmentation aims to accurately delineate object boundaries in an image according to a given textual prompt. Existing unsupervised or weakly supervised methods often rely on network-level annotated datasets and specially designed models to achieve precise segmentation. With the development of Latent Diffusion Models (LDMs)~\cite{pnvr2023ld}, impressive results have been demonstrated in text-to-image generation. Prior studies ~\cite{pnvr2023ld} have shown that LDMs inherently encode rich instance-level text–image alignment, which has sparked growing interest in extending their use beyond generation tasks toward semantic segmentation.

% For segmentation tasks based on diffusion models, the prevailing approach ~\cite{pnvr2023ld} involves freezing the diffusion-related modules to serve as a feature extractor, with a final head layer producing the segmentation output. VPD~\cite{zhao2023unleashing} leverages both attention features and trainable adapters and ADPP~\cite{pang2025aligning} conducting multi-round interaction in an agentic workflow to better align visual contents with textual prompts. While this strategy enables the model to quickly acquire multimodal knowledge and demonstrates strong generalization, it often yields coarse masks with imprecise boundaries. This limitation arises from a fundamental mismatch between the generative nature of diffusion processes and the discriminative nature of image segmentation. Generative modeling emphasizes diversity, where a single input may correspond to multiple acceptable outputs. In contrast, segmentation requires determinism, embodying a one-to-one mapping where a specific image and query must correspond to a single, well-defined mask.

For segmentation tasks based on diffusion models, LD-ZNet~\cite{pnvr2023ld} freezes diffusion-related modules and uses them as feature extractors, with a task-specific head producing the segmentation output. VPD~\cite{zhao2023unleashing} leverages attention features together with trainable adapters to improve the alignment between visual contents and textual prompts. ADPP~\cite{pang2025aligning} investigates the alignment between the generative denoising process and discriminative perception objectives, and exploits the denoising process as a controllable interface to support multi-round interactions within an agentic workflow for text-guided segmentation. While these approaches enable models to quickly acquire multimodal knowledge and demonstrate strong generalization, it often yields coarse masks with imprecise boundaries. This limitation arises from a fundamental mismatch between the generative nature of diffusion processes and the discriminative nature of image segmentation. Generative modeling emphasizes diversity, where a single input may correspond to multiple acceptable outputs. In contrast, segmentation requires determinism, embodying a one-to-one mapping where a specific image and query must correspond to a single, well-defined mask.

Rectified Flow (RF) addresses this challenge by learning a deterministic, near-linear Ordinary Differential Equation (ODE) trajectory between the source and target domains. This property aligns naturally with the requirements of image segmentation, making RF particularly well-suited for cross-task adaptation. As a result, it establishes a solid theoretical foundation for using diffusion models as backbones, enabling a smoother and more principled transition from generative to discriminative tasks. While SemFlow~\cite{wang2024semflow} introduces a bidirectional mapping between unconditional segmentation and conditional image generation, it remains limited in cross-modal modeling, as it does not incorporate textual guidance, and its applicability is confined to predefined semantic categories.
% RF fundamentally simplifies the process of adapting diffusion models for image segmentation. Firstly, it reformulates the learning objective into a direct image-to-mask trajectory, thus eliminating the need for the noise-injection process typical in generative training. Secondly, and of equal importance for the task transfer of generative models like Stable Diffusion, we circumvent the challenging problem of selecting an optimal timestep. These advantages significantly reduce the cost and complexity of task migration, making the adaptation of large pre-trained models more efficient and straightforward.

\begin{figure}[t]
    \centering
    \includegraphics[width=0.95\linewidth]{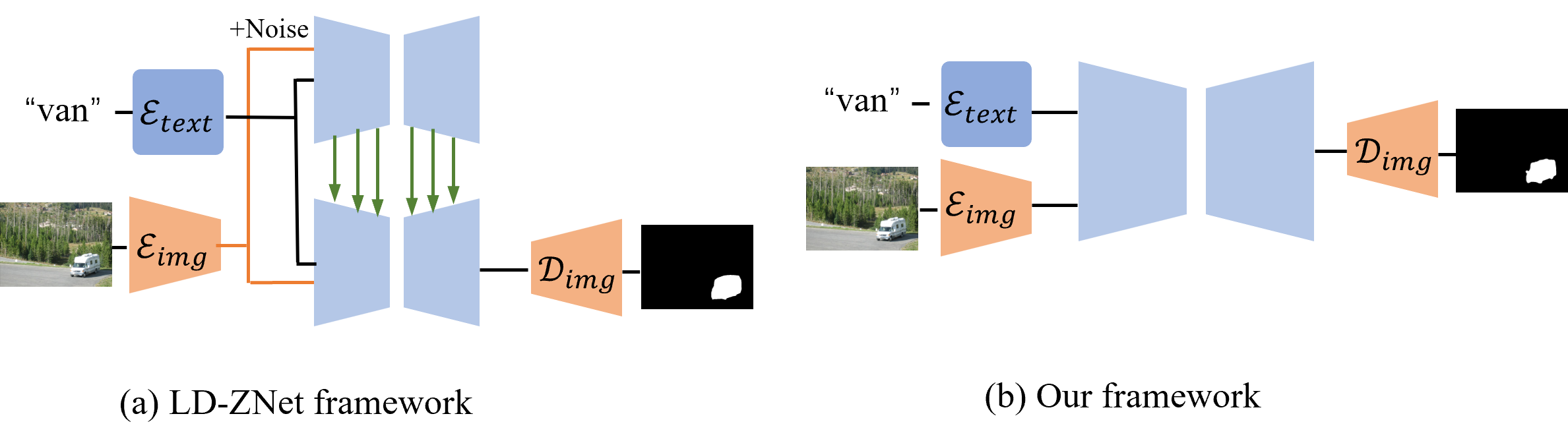}
    \caption{Prior methods rely on LDM as a feature extractor with extra branches, while ours directly enables text-based segmentation via finetuning.}
    \label{comparison_with_ld_znet_pipeline}
    % \vspace{-1.5em}
    \vskip -2em
\end{figure}

Based on the observations above, we propose a novel framework, named RLFSeg, that leverages the strengths of Latent Diffusion Models (LDMs) to generate high-quality and precise segmentation masks. Our framework consists of three key components: Rectified Latent Flow, Refinement and Dynamic Selection, and Adaptive One-Step Sampling. Existing methods often rely on stepwise denoising or additional UNet branches to extract latent features for mask generation, which incurs extra computational cost and risks error propagation. As shown in Figure~\ref{comparison_with_ld_znet_pipeline}, unlike previous approaches that use the model as a backbone, Rectified Latent Flow directly transforms image latents to mask latents in a single step, efficiently capturing semantic guidance from textual prompts. To further enhance mask quality and reduce the impact of annotation noise, we introduce a refinement and dynamic selection module that iteratively sharpens object boundaries and adaptively alternates between original and refined labels during training. Finally, our adaptive one-step sampling mechanism dynamically scales the latent update to ensure accurate boundary coverage within a single-step sampling process. In contrast to SemFlow~\cite{wang2024semflow}, our method focuses exclusively on text-driven segmentation. Through lightweight fine-tuning, it supports arbitrary textual and referring inputs, and explicitly optimizes for segmentation precision. By integrating these components, our framework produces segmentation masks that are both semantically aligned with the text input and visually precise, while remaining computationally efficient and robust to noisy annotations.

Extensive experiments demonstrate the effectiveness of our proposed framework. Our method achieves state-of-the-art results on multiple text-to-image segmentation benchmarks, including PhraseCut~\cite{wu2020phrasecut}, RefCOCO~\cite{kazemzadeh2014referitgame}, RefCOCO+~\cite{kazemzadeh2014referitgame}, and G-Ref~\cite{nagaraja2016modeling}. In summary, the main contributions of this work are as follows:
\begin{itemize}
\item We introduce Rectified Latent Flow to reconcile the generative nature of diffusion with the deterministic demands of segmentation by learning a direct, latent image-to-mask transformation.
\item We introduce label Refinment and Dynamic Selection (RDS) module to iteratively improve mask quality and mitigate annotation noise.
\item We design an Adaptive One-step Sampling (AOS) mechanism, improving boundary accuracy and overall mask precision in a single step.
\item Extensive experiments demonstrate that our method achieves state-of-the-art performance in both mIoU and AP.
\end{itemize}
In the spirit of transparency, we state that the Gemini model was utilized to refine the phrasing of this paper for enhanced readability. 

\section{RELATED WORK}
\subsection{Text-based Image Segmentation}
Text-based image segmentation aims to create pixel-level masks from free-form text, offering flexibility beyond fixed categories by handling both ``stuff'' and ``instances''. The field has evolved through several paradigms. Early works fused features from RNN and CNN backbones~\cite{hu2016segmentation,li2018referring,shi2018key,ye2019cross}, later enhanced by attention mechanisms for better cross-modal alignment~\cite{margffoy2018dynamic,wang2022cris,yu2018mattnet}. A significant shift occurred with large-scale models like CLIP~\cite{radford2021learning}, which improved representation learning and led to powerful foundation models such as SAM~\cite{kirillov2023segment} and SEEM~\cite{zou2023segment} with strong zero-shot capabilities. More recently, the trend has moved towards integrating segmentation into Vision-Language Large Models (VLLMs) for conversational reasoning. These models evolved from coarse bounding box grounding~\cite{chen2023shikra,you2023ferret} to direct mask prediction~\cite{lai2024lisa,ren2024pixellm,rasheed2024glamm}. However, despite their progress, these discriminative approaches often exhibit significant limitations. Many struggle to generate highly precise boundaries for complex, free-form instructions, while others require complex architectural modifications and costly fine-tuning to adapt to new tasks. This motivates exploring alternative generative paradigms, which may offer a more principled and effective approach to this task.

\subsection{Text-to-Image Synthesis}
% Text-to-image (T2I) synthesis has advanced rapidly from early GAN-based~\cite{xu2018attngan,zhu2019dm,tao2022df,zhang2021cross,ye2021improving,zhou2022towards} and autoregressive~\cite{ramesh2021zero,ding2021cogview,gafni2022make} models with vector-quantized autoencoders~\cite{van2017neural,razavi2019generating,esser2021taming} to diffusion models~\cite{nichol2021improved,dhariwal2021diffusion}, which significantly improve image quality and diversity. Pixel-space diffusion models are computationally expensive, motivating latent diffusion models (LDMs)~\cite{nichol2021glide,gu2022vector,tang2022improved,rombach2022high} that enable efficient high-resolution synthesis. Large-scale models such as Stable Diffusion~\cite{esser2024scaling,podell2023sdxl} and Imagen~\cite{baldridge2024imagen}, along with recent transformer-based architectures~\cite{flux,podell2023sdxl,peebles2023scalable}, further enhance photorealism and scalability. These developments establish diffusion models as a dominant T2I paradigm and provide rich semantic features useful for downstream tasks such as text-based image segmentation.
Text-to-image (T2I) synthesis has advanced rapidly from early GAN-based~\cite{xu2018attngan,zhu2019dm,tao2022df,zhang2021cross,ye2021improving,zhou2022towards} and autoregressive~\cite{ramesh2021zero,ding2021cogview,gafni2022make} models with vector-quantized autoencoders~\cite{van2017neural,razavi2019generating,esser2021taming} to diffusion models~\cite{nichol2021improved,dhariwal2021diffusion}, which achieve substantial improvements in image quality, diversity, and training stability. Despite their strong generation capability, early pixel-space diffusion models incur prohibitive computational costs, motivating the development of latent diffusion models (LDMs)~\cite{nichol2021glide,gu2022vector,tang2022improved,rombach2022high} that perform diffusion in a compact latent space and enable efficient high-resolution image synthesis. Building upon LDMs, large-scale models such as Stable Diffusion~\cite{esser2024scaling,podell2023sdxl} and Imagen~\cite{baldridge2024imagen}, together with recent transformer-based architectures~\cite{flux,podell2023sdxl,peebles2023scalable}, further improve photorealism, semantic consistency, and scalability across diverse visual concepts. Beyond image generation, these diffusion models are shown to encode rich text–image semantic correspondences, establishing them as a dominant paradigm for T2I synthesis and providing strong representational foundations for downstream tasks, including text-based image segmentation.

\subsection{Generative Models for Text-based Segmentation}
The remarkable scalability and transferability of diffusion models~\cite{nichol2021improved} make them a promising foundation for segmentation. Early work showed that features from generative models could be repurposed for this task~\cite{baranchuk2021label}, though often in limited few-shot~\cite{fei2006one} or domain-specific settings~\cite{karras2019style,yu2015lsun}. More recently, diffusion models have been adapted for text-driven segmentation via two main strategies. Training-free methods~\cite{corradini2024freeseg,karazija2023diffusion} align internal features with text but yield coarse boundaries, as the features are optimized for generation. To improve precision, training-based adaptations are used, but they often introduce significant overhead through complex multi-stage pipelines~\cite{li2023open}, auxiliary modules, or costly alignment training~\cite{pnvr2023ld,stracke2025cleandift}. These strategies treat the diffusion model as a component rather than reframing its core process for segmentation. In contrast, our method offers a more fundamental solution by employing Rectified Flow to reframe the task. We directly fine-tune a pretrained LDM to learn a deterministic, single-step mapping from image to mask, effectively transforming the stochastic, multi-step generation process and achieving superior segmentation performance.

% Training of Diffusion-based generative models~\cite{nichol2021improved} are robust and scalable, and features from pretrained diffusion models can be repurposed for semantic segmentation~\cite{baranchuk2021label}, albeit mostly in few-shot~\cite{fei2006one} or domain-specific~\cite{karras2019style,yu2015lsun} settings. Leveraging their rich language–vision correspondences, diffusion models have been applied to text-driven segmentation: some methods~\cite{li2023open} use Stable Diffusion with external detectors to generate pseudo-labeled datasets, while training-free approaches~\cite{corradini2024freeseg,karazija2023diffusion} align diffusion features with text for zero-shot mask prediction, though the output boundaries remain coarse. 
% Other works~\cite{pnvr2023ld,stracke2025cleandift} respectively require an extra module and alignment training. In contrast, our method directly fine-tunes a pretrained LDM in latent space, learning a rectified flow from image to mask distribution in a single step, enabling faster inference and more precise, semantically aligned segmentation.

\section{METHOD}
\begin{figure*}[t]
    \centering
    \includegraphics[width=0.98\linewidth]{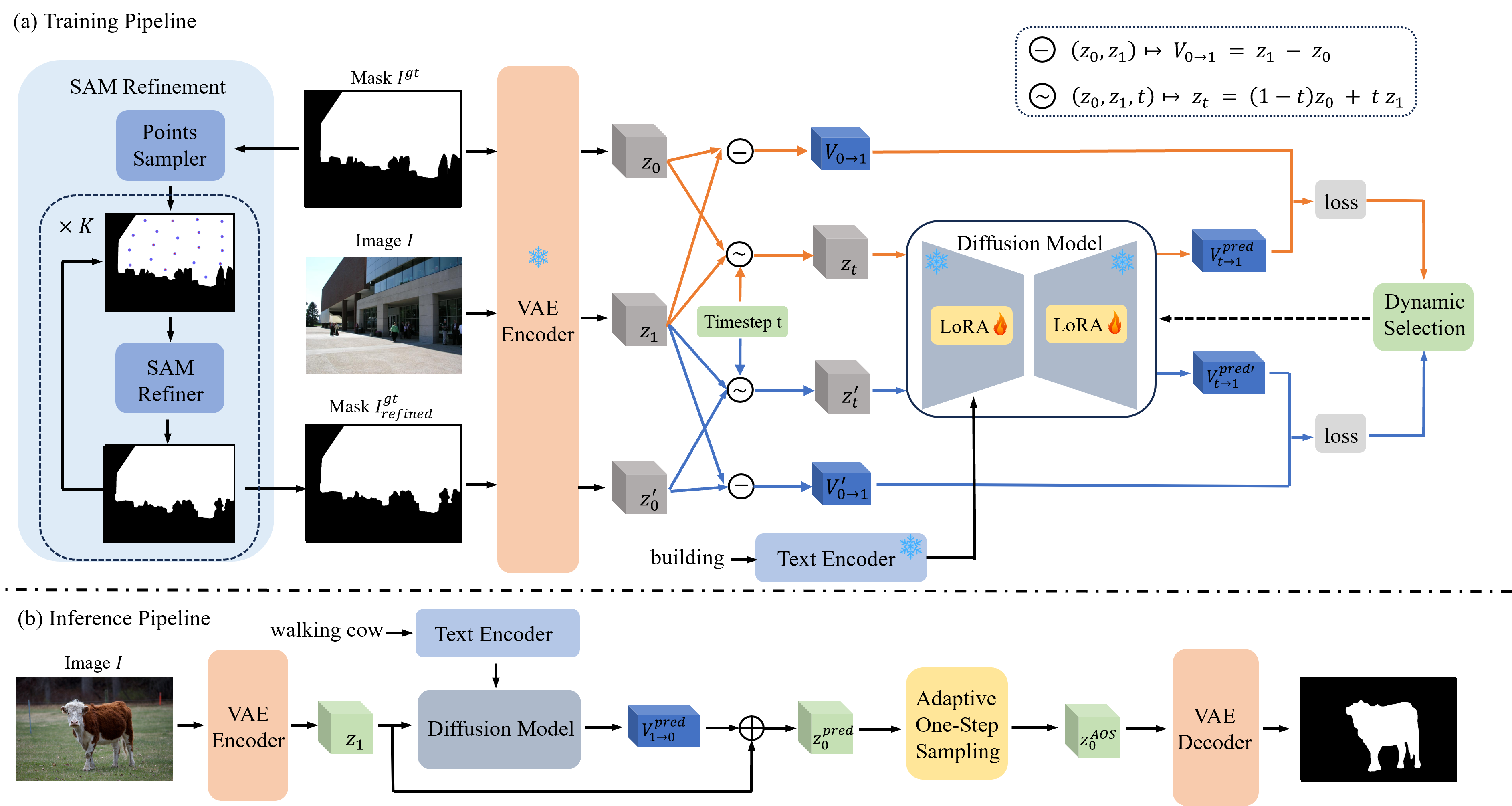}
    \caption{\textbf{Overview of RLFSeg.} (a) Training pipeline with Rectified Latent Flow and SAM-driven Label Refinement, where ground-truth labels are automatically matched with SAM-refined annotations for finer supervision by iteratively sampling points to progressively refine masks. (b) Inference pipeline with Adaptive One-Step Sampling to dynamically adjust the step size for sampling.}
    \label{pipeline}
    \vspace{-0.5em}
\end{figure*}

In this section, we first introduce the preliminary knowledge required for understanding the key components of our method. We then detail our proposed framework, RLFSeg, which enhances text-to-image segmentation by leveraging the strengths of Latent Diffusion Models (LDMs). Our method consists of three core components: 1) Rectified Latent Flow, which refines latent flow to directly generate segmentation masks from the original image in a single step; 2) Refinement and Dynamic Selection(RDS), which utilizes the Segment Anything Model (SAM)~\cite{kirillov2023segment} for label optimization and automatic loss selection; 3) Adaptive One-Step Sampling(AOS), which dynamically adjusts the norm of the predicted velocity \(v\), enabling our approach to achieve high-quality results in a single sampling step. The pipeline of our method is illustrated in Figure~\ref{pipeline}.

\subsection{Preliminaries}
\textbf{Latent Diffusion Models}(LDMs)~\cite{rombach2022high} generate images in a compressed latent space through two main stages. First, an autoencoder (e.g., {VQGAN~\cite{esser2021taming})} maps an input image $x$ to a latent representation $z= \Phi_{encoder}(x)$, preserving its semantic content in a compact form. Then, a diffusion UNet learns to iteratively denoise the latent via a reverse process, guided by text features extracted from a pretrained CLIP encoder~\cite{radford2021learning} through cross-attention. The denoising process can be written as
\begin{equation}
z_t = f_{\theta}(z_{t-1}, t, c),
\label{eq2}
\end{equation}
where $z_t$ is the latent at step $t$, $f_\theta$ is the denoising function, and $c$ denotes the text condition. This formulation enables efficient and high-quality text-to-image generation.

\subsection{Rectified Latent Flow}

The training paradigm of diffusion models, such as DDPM, is centered around a progressive noising process on target data to construct a generative path from pure noise to real data. This stochastic perturbation mechanism is the fundamental reason for the diversity in the outputs of diffusion models. However, an inherent task conflict arises when this paradigm is directly applied to image segmentation. The conventional practice involves conditioning on the source image while modeling the noised mask, which essentially forces a discriminative task that seeks a unique, deterministic solution into a generative framework designed for diversity. For segmentation, a given image and text prompt should map to a single, deterministic mask. Therefore, noising the mask to imitate the generation process is a circuitous and unnecessary redundant design.

The emergence of Rectified Flow provides an elegant solution to this cross-task alignment dilemma. As a new-generation generative paradigm, it aims to learn a deterministic, near-linear Ordinary Differential Equation (ODE) path between source and target domains, which is theoretically consistent with the optimal transport (OT) path connecting the image and the mask. This mechanism obviates the need for stochastic noising. Given the inherent compatibility between the determinism of Rectified Flow and the intrinsic demands of segmentation, we recognized that it provides a solid theoretical bridge for leveraging the powerful pre-trained knowledge of diffusion models. To this end, we propose RLFSeg, a method designed to directly learn a continuous, direct mapping path from the source image to the target mask, thereby seamlessly aligning the powerful generalization capabilities of the generative model with the objectives of the discriminative task.

Given an input image \( I \) and a mask \( M \), we first use the VAE encoder \( \Phi_{encoder} \) to obtain their latent representations, \( z_1 = \Phi_{encoder}(I) \) and \( z_0 = \Phi_{encoder}(M) \). We then extract frozen CLIP text features from the provided text prompt, which are fed into the denoising UNet of the LDM. This setup allows the model to generate segmentation masks by leveraging semantic guidance from the text input.

During training, we directly learn the vector field \( \mathbf{v} = z_1 - z_0 \) that defines the straight path between the image latent \(z_0\) and the mask latent \(z_1\).Following Rectified Flow, we sample a time \(t\) from a uniform distribution \(U(0, 1)\) and construct an intermediate latent \(z_t=tz_1+(1-t)z_0\). In summary, the Rectified Flow loss is defined in Eq.\ref{rf_loss}:
\begin{equation}
\label{rf_loss}
\mathcal{L}_{rf}=\min_{v} \int_{0}^{1} E\left[\left\|(z_{1} - z_{0}) - v_\theta(z_{t}, t)\right\|^{2}\right] dt,
\end{equation}
where \(v_\theta\) is the model which is trained to predict the constant vector field \(\mathbf{v}\) given the interpolated latent \(z_t\) and timestep \(t\) .

\subsection{Refinement with  dynamic selection}

\begin{figure}[t]
    \centering
    \includegraphics[width=0.98\linewidth]{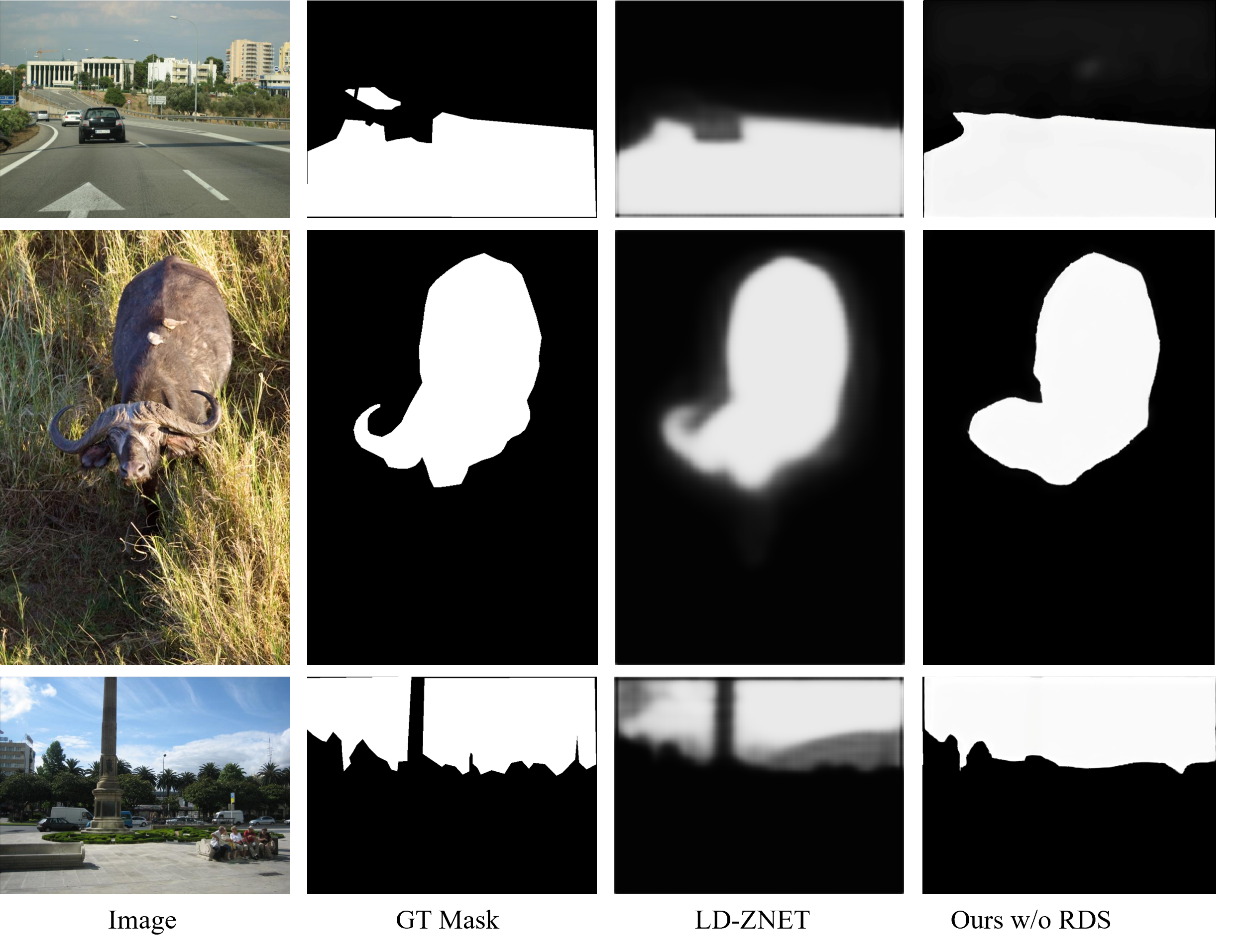}
    \caption{\textbf{Visualization of results without RDS.} The Rectified Flow training can cause the model to predict polygon-like masks in some cases, which reduces segmentation accuracy.}
    \label{effect_polyon_anno}
    \vspace{-0.5em}
\end{figure}

Segmentation masks in our dataset are generated from polygon-based annotations, which are often coarse and prone to noise, limiting their effectiveness for precise text-based segmentation. This issue is particularly exacerbated by the training methodology of Rectified Flow, as it makes the model more susceptible to learning the annotation style of the dataset. Figure~\ref{effect_polyon_anno} showcases several illustrative examples from the test set where this stylistic overfitting is apparent. To address this, we propose a SAM-driven Label Refinement and Dynamic Selection Module that iteratively improves mask quality and adaptively leverages both original and refined labels during training. 

\textbf{SAM-driven Label Refinement} iteratively applies SAM to refine the boundaries of segmentation masks. To provide the SAM model with guiding prompt, we generated a set of N points \( P \) by applying k-means clustering to the initial mask via Eq.~\ref{kmeans} . Notably, these points remain fixed throughout the iterative process, thus serving as spatial anchors that effectively reduce the cumulative positional drift of the mask. The algorithm is demonstrated in Algorithm \ref{alg:refine}, where \(\tau\) is set to 0.99.

% To satisfy the point prompt requirement of the SAM, we generate a set of N points \( P \) by applying k-means clustering to the initial mask via Equation \ref{kmeans} . This strategy serves a dual purpose: not only does it provide the necessary spatial guidance for SAM, but more importantly, since these points remain constant throughout the iterative process, they act as a set of stable spatial anchors, effectively mitigating the cumulative drift of the mask's position.

\begin{equation}
\label{kmeans}
    \mathbf{P} = \text{KMeans}\left( \{ (x, y) \mid \mathbf{M}_{0}(x, y) > 0.5 \}, N \right) ,
\end{equation}
where \(\mathbf{M}_{0}(x, y)\) represents the value of the pixel at coordinates \((x,y)\) in the ground-truth mask.

In each iteration, SAM (denoted by \( \Phi_{SAM}\)) predicts a new mask \(M_t\) based on the previous mask \(M_{t-1}\) and the anchor points \(P\). The process terminates once the IoU score converges. Through this process, the mask boundary is progressively refined to more accurately match the object's contour.

% \begin{algorithm}[H]
% \caption{Iterative Mask Refinement}
% \label{alg:refine}
% \begin{algorithmic}[1]
%     \Require Initial mask $M_0$, Prompt $P$, Max iterations $T_{max}$
%     \Ensure Refined mask $M_{\text{refine}}$
%     \For{$t=1$ \to $T_{max}$}
%         \State $M_t \gets \Phi_{\text{SAM}}(P, M_{t-1})$,  $\text{IoU}_t \gets \operatorname{IoU}(\mathbf{M}_t, \mathbf{M}_{t-1})$
%     \EndFor
% \State $t_{\text{best}} \gets \arg\min_t \text{IoU}_t$, $\mathbf{M}_{\text{refine}} \gets \mathbf{M}_{t_{\text{best}}}$
% \end{algorithmic}
% \end{algorithm}

% \begin{algorithm}[H]
% \caption{Iterative Mask Refinement with Early Stopping}
% \label{alg:refine}
% \begin{algorithmic}[1]
%     \Require Initial mask $M_0$, Prompt $P$, Max iterations $T_{max}$, Threshold $\tau$
%     \Ensure Refined mask $M_{\text{refine}}$
%     \For{$t=1$ \to $T_{max}$}
%         \State $M_t \gets \Phi_{\text{SAM}}(P, M_{t-1})$,  $\text{IoU}_t \gets \operatorname{IoU}(\mathbf{M}_t, \mathbf{M}_{t-1})$
%         \If{$\text{IoU}_t < \tau$}
%             \State \textbf{break}
%         \EndIf
%     \EndFor
% \State $\mathbf{M}_{\text{refine}} \gets \mathbf{M}_{t}$
% \end{algorithmic}
% \end{algorithm}

\begin{algorithm}[H]
\caption{Iterative Mask Refinement with Early Stopping}
\label{alg:refine}
\begin{algorithmic}[1]
    \Require $M_0$, $P$, $T_{max}$, $\tau$
    \Ensure $M_{\text{refine}}$
    \For{$t=1 \to T_{max}$ \textbf{ until } $\operatorname{IoU}(M_t,M_{t-1})<\tau$}
        \State $M_t \gets \Phi_{\text{SAM}}(P, M_{t-1})$
    \EndFor
    \State $M_{\text{refine}} \gets M_t$
\end{algorithmic}
\end{algorithm}

\textbf{Dynamic Selection}, as defined in Eq.~\ref{LDS}, automatically selects between the losses computed on the original mask and the refined mask. This strikes a crucial balance between the stability of the original annotations and the precision of the refined results. As our ablation studies in Section~\ref{ablation_study} will demonstrate, this approach yields significant performance gains.

\begin{equation}
\label{LDS}
\mathcal{L}_{final}=\mathbf{1}_{rds}\mathcal{L}_{rf}(z_0=\mathbf{M}_0)+(1-\mathbf{1}_{rds})\mathcal{L}_{rf}(z_0=\mathbf{M}_{refine}) ,
\end{equation}
where \( \mathbf{1} \) is an indicator function that equals 1 if \(\mathcal{L}_{rf}(z_0=\mathbf{M}_0) < \mathcal{L}_{rf}(z_0=\mathbf{M}_{refine})\), and 0 otherwise. The design follows a core logic: a higher-quality annotation simplifies the learning task, providing a clearer path for the model to converge to a lower loss. Although this scenario is relatively uncommon (occurring in approximately \(15\%\) of our training cases), when the refined annotation \(M_{refine}\) paradoxically degrades in quality, the model will instead learn from the original label \(M_0\).

\begin{figure*}[t]
    \centering
    \includegraphics[width=0.9\linewidth]{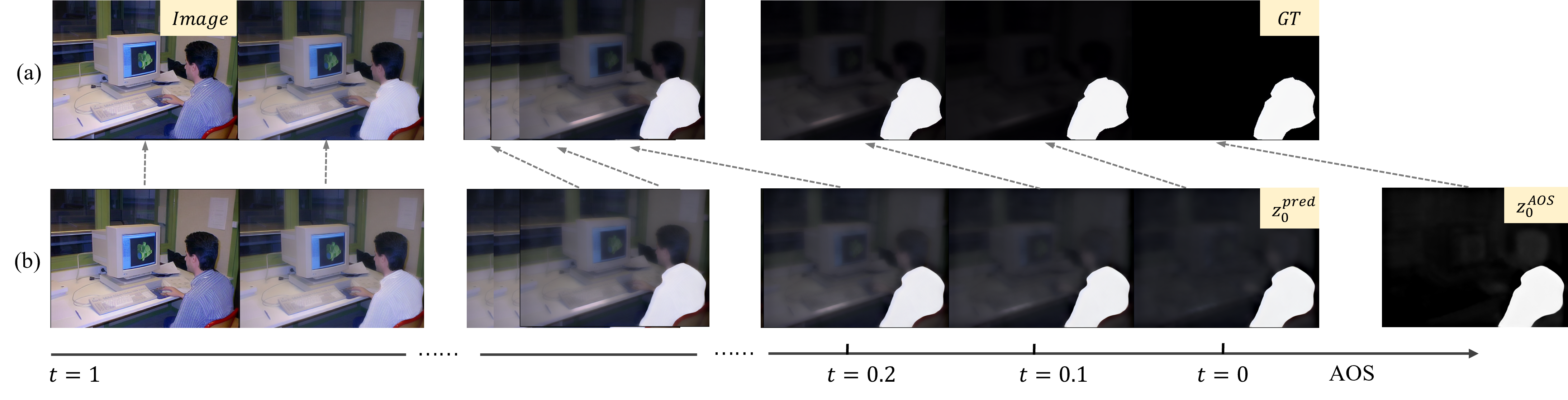}
    \caption{\textbf{Comparison of sampling trajectories and the effect of AOS.} 
    Interpolation with (\textbf{a}) the ground-truth \(z_0^{truth}\) versus (\textbf{b}) our prediction \(z_0^{pred}\).  The artifact-laden \(z_0^{pred}\) misaligns with the final target \(z_0^{truth}\) , corresponding instead to an intermediate ground truth \(z_t^{truth}\) (e.g., \(t=0.1\)). Our AOS corrects this misalignment by adaptively scaling the update step, yielding a much sharper and more accurate result.}
    \label{fig:trajectories}
    \vspace{-1.5em}
\end{figure*}

\subsection{ADAPTIVE ONE-STEP SAMPLING}

Motivated by the path-crossing problem in multi-step sampling (Experiment~\ref{path_crossing}), we find that single-step sampling via  \(z_0^{pred}=z_1+\mathbf{v}_{1 \to 0}\) yields comparable or superior performance. Nevertheless, we have observed that for a subset of challenging samples, this single-step process can yield blurry predictions. Based on Figure~\ref{fig:trajectories}, these artifacts resemble an intermediate state along the flow trajectory that has not fully converged to the target. We hypothesize that this phenomenon is attributable to the model's underestimation of the magnitude of the predicted velocity vector \( \mathbf{v} \) . Consequently, the predicted state \(z_0^{pred}\) fails to completely reach the ground-truth target \(z_0^{truth}\), thereby introducing a residual offset between them. To rectify this issue, we propose a novel methodology termed Adaptive Single-Step Sampling.

To ensure precise alignment of \(z_0^{\text{pred}}\) with the ground-truth distribution, AOS anchors the background regions of the predicted latent to a black reference latent \(z_b = \Phi_{\text{encoder}}(I_b)\), where \(I_b\) is a pure black image. Given the predicted velocity
\begin{equation}
\Delta z_{1 \to 0}^{\text{pred}} = \mathbf{v}_{1 \to 0}^{\text{pred}} \cdot \Delta t, 
\quad z_0^{\text{pred}} = z_1 + \Delta z_{1 \to 0}^{\text{pred}},
\end{equation}
we first compute a candidate latent \(z_0^{\text{pred}}\) from the single-step update. To identify reliable points for correction, we introduce the stable region index set
\begin{equation}
\mathcal{S} = \big\{ i \;\big|\; \big| z_b[i] - z_0^{\text{pred}}[i] \big| < \epsilon \big\},
\end{equation}
where \(\epsilon=0.01\). This set collects the positions where the predicted latent is sufficiently close to the black reference latent. These points are assumed to correspond to background areas that should remain consistent across updates.  

Within this stable region \(\mathcal{S}\), we measure the average deviation between \(z_0^{\text{pred}}\) and the reference latent relative to the predicted update magnitude. This ratio defines the adaptive scaling factor:
\begin{equation}
\gamma = 
\frac{ \sum_{i \in \mathcal{S}} \big| z_b[i] - z_0^{\text{pred}}[i] \big|}
{ \sum_{i \in \mathcal{S}} \big| \Delta z_{1 \to 0}^{\text{pred}}[i] \big|},
\qquad
z_0^{\text{AOS}} = z_1 + \Delta z_{1 \to 0}^{\text{pred}} \cdot (1 + \gamma).
\end{equation}

By adaptively rescaling the update in proportion to the deviation observed in stable background regions, AOS compensates for prediction drift and enforces more consistent background alignment. This mechanism improves the overall trajectory of the rectified flow, leading to more accurate boundary localization and greater robustness against noisy predictions.

\section{EXPERIMENTS}
\subsection{Experiments Setting}
\textbf{Implementation details.} In our experiments, we adopt the Stable Diffusion v1.5~\cite{rombach2022high} checkpoint as the foundation for our Latent Diffusion Model (LDM), which utilizes the ViT-L/14 CLIP text encoder~\cite{radford2021learning} in a frozen state. During the training phase, we preserve the original parameters of the Stable Diffusion model and focus on fine-tuning only the LoRA~\cite{hu2022lora} layers, with a fixed rank of 64 applied throughout. The training is conducted on 8 NVIDIA A100 GPUs, each processing a batch size of 8, with the Adam optimizer and a base learning rate of 1e-4 for each mini-batch sample on each GPU.

\textbf{Dataset.} We follow LD-ZNet~\cite{pnvr2023ld} and evaluate on several benchmark datasets. PhraseCut~\cite{wu2020phrasecut}, the largest dataset for text-based image segmentation with ~340K phrase–mask pairs, provides annotations for both stuff classes and multiple object instances. To further test generalization, we take the model trained on the PhraseCut training set and directly evaluate it on the referring expression segmentation benchmarks RefCOCO~\cite{kazemzadeh2014referitgame}, RefCOCO+~\cite{kazemzadeh2014referitgame}, and G-Ref~\cite{nagaraja2016modeling}. RefCOCO consists of short expressions (avg. 3.6 words) with at least two objects per image, while RefCOCO+ removes location words and focuses on appearance-based descriptions, making it more challenging. G-Ref contains longer expressions (avg. 8.4 words) with richer appearance and location details. We adopt the UNC partition for RefCOCO/RefCOCO+ and the UMD partition for G-Ref.

\textbf{Metrics.} Following LD-ZNet~\cite{pnvr2023ld}, we report two evaluation metrics: the best mean Intersection-over-Union (mIoU) and the Average Precision (AP). The mIoU measures the overall pixel-level overlap between the predicted segmentation and the ground truth, providing a comprehensive assessment of segmentation accuracy. The AP evaluates the precision–recall trade-off across different thresholds, reflecting the model’s ability to localize the regions referred to in text.

\begin{table}[htbp]
\centering
\small
\caption{Performance comparison of text-based segmentation methods on the PhraseCut test set. Our approach outperforms all other compared methods in terms of mIoU.}
\label{tab:segmentation_results}
\begin{tabular}{lcc}
\toprule
Method & mIoU & AP \\
\midrule
RMI & 21.1 & - \\ 
Mask-RCNN Top & 39.4 & - \\
HulaNet & 41.3  & - \\
CLIPSeg (PC+) & 43.4  & 76.7 \\
CLIPSeg (PC, D=128)& 48.2 & 78.2 \\
RGBNet & 46.7 & 77.2 \\
LD-ZNet & 52.7 & \textbf{78.9} \\
RLFSeg(Ours) &\textbf{56.1} & 77.3\\
\bottomrule
\end{tabular}
\end{table}

\begin{figure*}[t]
     \includegraphics[width=0.9\linewidth]{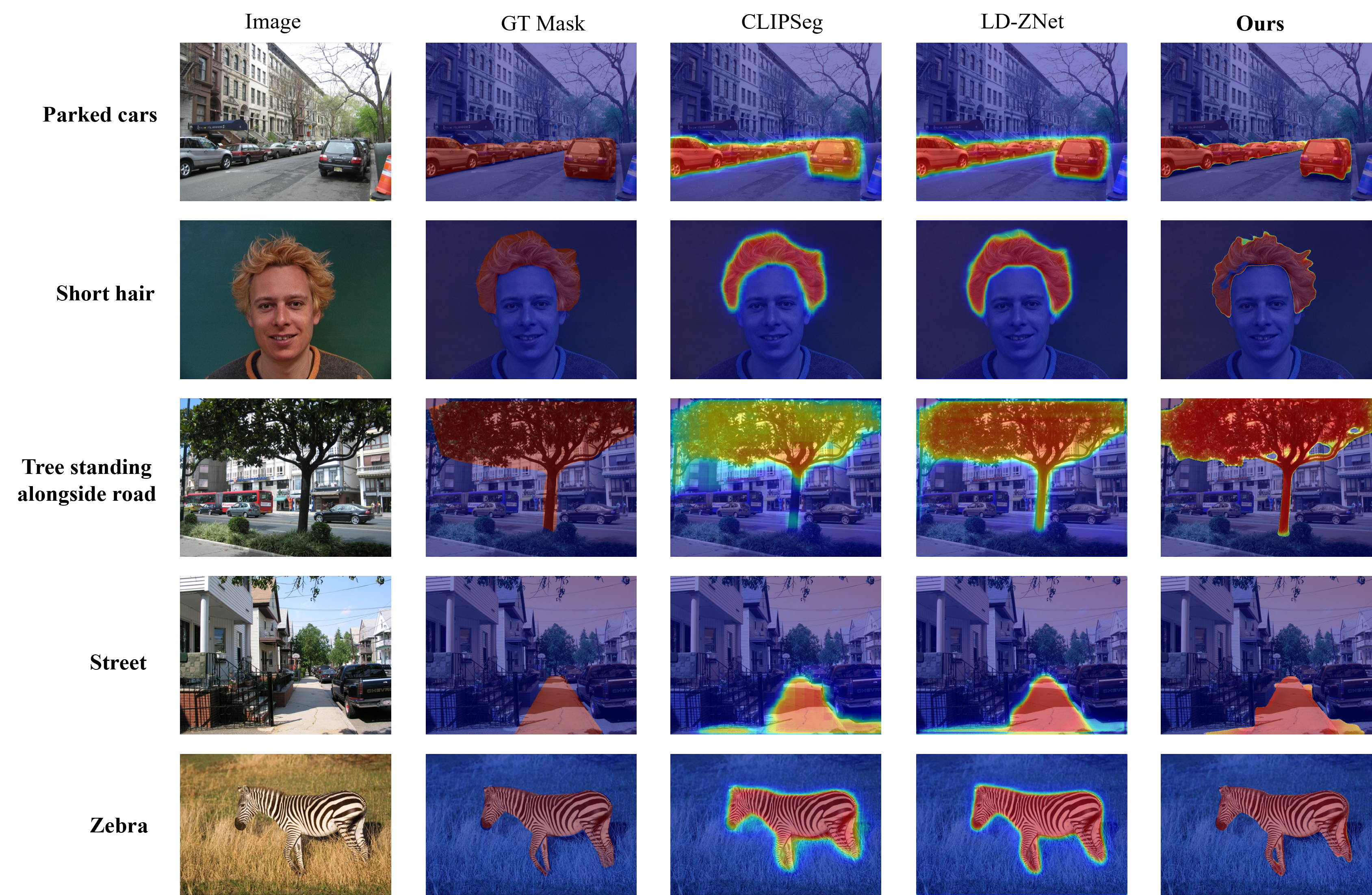}
    \captionof{figure}{\textbf{Qualitative comparison with different methods.}}
    \label{compare}
    % \vspace{-1.5em} 
    \vskip -1em
\end{figure*}

\begin{table}[h]
\centering
\caption{Text-based segmentation results on RefCOCO, RefCOCO+, and G-Ref. We report mean IoU (mIoU) and Average Precision (AP) for each dataset. For fair comparison, all SD-based baseline models were pre-trained on the same dataset as Stable Diffusion v1.5.}
\label{tab:refexp_results}
\begin{tabular}{lcccccc}
\toprule
Method & \multicolumn{2}{c}{RefCOCO} & \multicolumn{2}{c}{RefCOCO+} & \multicolumn{2}{c}{G-Ref} \\
\cmidrule(lr){2-3} \cmidrule(lr){4-5} \cmidrule(lr){6-7}
 & mIoU & AP & mIoU & AP & mIoU & AP \\
\midrule
\multicolumn{7}{l}{\textit{Zero-shot}} \\ % 这里改成了 {c}
\midrule
CLIPSeg (PC+) & 30.1 & 14.1 & 30.3 & 15.5 & 33.8 & 23.7 \\
RGBNet & 36.3 & 15.7 & 37.1 & 16.7 & 41.9 & 27.8 \\
LD-ZNet & 41.0 & 17.2 & 42.5 & 18.6 & 47.8 & 30.8 \\
RLFSeg(Ours) &\textbf{42.5} &\textbf{50.9} &\textbf{43.4} &\textbf{52.5} &\textbf{51.8} &\textbf{60.0} \\
\midrule
\multicolumn{7}{l}{\textit{Fully Supervised}} \\ % 第二部分标题
\midrule
VPD & 73.3 & - & 62.7 & - & 62.0 & - \\
ADDP & 69.1 & - & 57.6 & - & 59.0 & - \\
RLFSeg(Ours) &\textbf{75.3} &\textbf{81.7} &\textbf{66.0} &\textbf{71.8} &\textbf{67.8} &\textbf{74.1} \\
\bottomrule
\vspace{-2em}
\end{tabular}
\end{table}

\begin{table}[htbp]
\centering
\small
\caption{Segmentation results on the COCO-Stuff dataset. Our method shows a notable improvement in mIoU.}
\label{tab:cocostuff_results}
\begin{tabular}{lcc}
\toprule
\multirow{2}{*}{\textbf{Method}} & \multicolumn{2}{c}{\textbf{COCO-Stuff}} \\
\cmidrule(lr){2-3}
& mIoU & AP \\
\midrule
SemFlow & 38.6 & - \\
\textbf{Ours} & \textbf{39.7} & \textbf{59.0} \\
\bottomrule
\end{tabular}
\vskip -2em
\end{table}

\subsection{Quantitative Evaluations}
% We conduct comprehensive quantitative comparisons on four widely used text-based segmentation benchmarks: PhraseCut~\cite{wu2020phrasecut}, RefCOCO~\cite{kazemzadeh2014referitgame}, RefCOCO+~\cite{kazemzadeh2014referitgame}, and G-Ref~\cite{nagaraja2016modeling}. The results are summarized in Tab.~\ref{tab:segmentation_results} and Tab.~\ref{tab:refexp_results}. 
We quantitatively compare our method, RLFSeg, with state-of-the-art and baseline methods on four standard benchmarks. Our evaluation includes prominent approaches such as the VDP\cite{zhao2023vpd}, the ADDP\cite{pang2025aligning}, the LD-ZNet~\cite{pnvr2023ld}, the CLIPSeg~\cite{luddecke2022image}, and other established methods like HulaNet~\cite{wu2020phrasecut} and RGBNet.

On the PhraseCut benchmark, our RLFSeg achieves the highest mIoU of 56.1 and a competitive AP of 77.3, surpassing previous methods such as CLIPSeg (48.2 mIoU)\cite{luddecke2022image}, RGBNet (46.7 mIoU), and LD-ZNet (52.7 mIoU), as detailed in Tab.\ref{tab:segmentation_results}. These results highlight RLFSeg’s ability to capture fine-grained semantics and generate precise segmentation masks directly from textual prompts. The phenomenon where our method achieves significantly higher mIoU (56.1 vs. 52.7) but slightly lower AP (77.3 vs. 78.9) on PhraseCut compared to LD-ZNet is expected. It stems from the fundamental difference between Generative Flows and Discriminative Classifiers. Discriminative baselines use Cross-Entropy loss to output smooth ambiguous boundaries, which benefits AP performance but weakens mIoU. Our Rectified Flow generative method generates sharp, deterministic boundaries for excellent mIoU. However, it narrows the threshold range for AP calculation, resulting in lower AP scores. Unlike prior methods that rely on extra U-Net branches (e.g.,LD-ZNet) or handcrafted architectural designs (e.g., RGBNet), RLFSeg leverages rectified latent flows with adaptive refinement, achieving more accurate boundary alignment without additional architectural complexity.

As detailed in Tab.~\ref{tab:refexp_results}, RLFSeg also consistently outperforms existing methods on the more challenging referring expression benchmarks. For instance, on RefCOCO, RLFSeg obtains 42.5 mIoU and 50.9 AP, far exceeding the strongest baseline, LD-ZNet (41.0 mIoU, 17.2 AP). On RefCOCO+, RLFSeg reaches 43.4 mIoU and 52.5 AP, again outperforming LD-ZNet (42.5 mIoU, 18.6 AP). The advantage is most pronounced on G-Ref, where RLFSeg achieves 51.8 mIoU and 60.0 AP, compared to 47.8 mIoU and 30.8 AP of LD-ZNet. These consistent improvements across diverse datasets highlight the strong zero-shot generalization ability of our framework in handling complex queries and challenging object boundaries.

We directly compare our method with SemFlow to highlight key performance and efficiency differences. As shown in Tab~\ref{tab:cocostuff_results}, our model achieves 39.7 mIoU on COCO-Stuff in a zero-shot setting, surpassing SemFlow's fully-trained result of 38.6 mIoU. This demonstrates the superior effectiveness of our architecture, which is optimized for discriminative precision. 

We detail the inference speed comparison of our proposed model against prior work in Tab.\ref{tab:inference_time}. Tested on an RTX 3090 (avg. 100 runs), the speed ranking is LD-ZNet  \(\gtrsim\)  Ours \(\gg\) SemFlow. We are only marginally slower than LD-ZNet due to the overhead of the standard SD VAE decoder (designed for RGB). Utilizing a lightweight, mask-specific VAE would eliminate this redundancy, allowing our method to surpass LD-ZNet.

\begin{table}[h]
\centering
\caption{Comparisons of Inference Time. The time is measured
in seconds per individual image, averaged over 100 runs.}
\label{tab:inference_time}
\begin{tabular}{lc}
\toprule
Method & Inference Time (s) \\
\midrule
LD-ZNet       & \textbf{0.17} \\
SemFlow       & 1.09 \\
\textbf{Ours} & 0.19 \\
\bottomrule
\end{tabular}
\end{table}

In summary, across all benchmarks, RLFSeg attains the best performance among diffusion-based segmentation approaches. By directly modeling latent transformations with rectified flows and incorporating refined supervision, our framework effectively narrows the generative–discriminative gap, producing segmentation masks that are both semantically consistent and boundary-accurate, while remaining efficient and robust to noisy annotations.

\subsection{Ablation Study}
\label{ablation_study}
We demonstrate the effectiveness of each component of RLFSeg through ablation studies, with the results presented in Tab.~\ref{tab:RDS_HSL}. Compared to the baseline model (trained solely with the RF strategy), which achieves 55.3 mIoU, introducing the RDS strategy during training refines the learning objective, leading to a more precise mapping and an improved mIoU of 55.8. Furthermore, applying AOS during the inference stage further refines the latent features predicted by RLFSeg, raising the mIoU to 56.1. Both strategies independently improve the performance of the RF model, and when combined, they provide cumulative performance gains.

We conducted an ablation study to determine the optimal LoRA rank for our fine-tuning process. As detailed in Tab.~\ref{lora_rank}, performance peaks at a rank of 64 (56.1 mIoU), with higher values yielding no significant improvement. Most strikingly, abandoning LoRA for full-parameter fine-tuning causes a severe performance collapse to 51.9 mIoU. This result strongly suggests that full fine-tuning catastrophically damages the model's essential pre-trained priors, underscoring that a parameter-efficient approach like LoRA is crucial for effectively leveraging the diffusion model for our task.

% Furthermore, Tab.~\ref{tab:steps_results} illustrates the performance variations across different sampling steps on multiple datasets. Notably, 1-step sampling consistently achieves the best mIoU results. This observation demonstrates that our AOS effectively corrects this misalignment by adaptively scaling the update step, resulting in substantially sharper and more accurate segmentation results.
Furthermore, Tab.~\ref{tab:steps_results} illustrates the performance variations across different sampling steps across multiple datasets. Notably, 1-step sampling consistently achieves the best mIoU results, while performance degrades as more sampling steps are introduced. This observation provides strong empirical evidence that our AOS effectively corrects the misalignment between the diffusion sampling process and the segmentation objective by adaptively scaling the update step, thereby preventing error accumulation at later denoising stages and yielding substantially sharper and more accurate segmentation results.

\begin{table}[htbp]
\centering
\small
\caption{Comparison of performance with different strategies. RDS denotes the Refinement with Dynamic Selection, and AOS denotes Adaptive One-Step Sampling.}
\label{tab:RDS_HSL}
\begin{tabular}{cc|cc}
\toprule
RDS & AOS & mIoU & AP \\
\midrule
\textcolor{red}{×} & \textcolor{red}{×} & 55.3 & 76.7 \\
\textcolor{red}{×} & \textcolor{green}{\checkmark} & 55.4 & \textbf{78.5} \\
\textcolor{green}{\checkmark} & \textcolor{red}{×} & 55.8 & 76.9 \\
\textcolor{green}{\checkmark} & \textcolor{green}{\checkmark} & \textbf{56.1} & 77.3 \\
\bottomrule
\end{tabular}
\end{table}

\begin{table}[htbp]
\centering
\small
\caption{Ablation study on the effect of LoRA rank on text-based segmentation performance. ``Full Params'' indicates the baseline model without LoRA.}
\label{lora_rank}
\begin{tabular}{c|cc}
\toprule
Rank & mIoU & AP \\
\midrule
16   & 55.3 & 76.7 \\
32   & 55.7 & 76.3 \\
64   & 56.1 & 77.0 \\
128  & 56.1 & 76.6 \\
256  & 56.0 & 77.5 \\
Full Params & 51.9 & 73.1 \\
\bottomrule
\end{tabular}
\vskip -1em
\end{table}

\begin{table}[h]
\centering
\caption{Comparison of Different Sampling Steps.}
\label{tab:steps_results}
{
\renewcommand{\arraystretch}{1.15}
\begin{tabular}{c cc cc cc cc}
\toprule
\multirow{2}{*}{Steps} &
\multicolumn{2}{c}{PhraseCut} &
\multicolumn{2}{c}{RefCOCO} &
\multicolumn{2}{c}{RefCOCO+} &
\multicolumn{2}{c}{G-Ref} \\
\cmidrule(lr){2-3}\cmidrule(lr){4-5}\cmidrule(lr){6-7}\cmidrule(lr){8-9}
& mIoU & AP & mIoU & AP & mIoU & AP & mIoU & AP \\
\midrule
\textbf{1} & \textbf{55.8} & \textbf{77.1} & \textbf{42.2} & \textbf{49.3} & \textbf{43.1} & \textbf{51.0} & \textbf{51.6} & \textbf{59.3} \\
2  & 47.2 & 72.7 & 37.0 & 44.2 & 36.7 & 45.2 & 44.2 & 53.8 \\
5  & 46.0 & 71.1 & 36.4 & 43.5 & 36.0 & 44.3 & 43.6 & 52.6 \\
15 & 44.5 & 69.7 & 35.6 & 43.0 & 35.2 & 43.8 & 42.7 & 51.8 \\
\bottomrule
% \vspace{-0.5em}
\end{tabular}
}
\end{table}

\subsection{Qualitative Analyses}
\label{subsec:qual_analyses} 
\textbf{Precision in Detail.} Figure~\ref{compare} showcases several examples where our method achieves more precise segmentation results. For instance, in the ``short hair'' case, our approach demonstrates a more accurate grasp of the hair's contours compared to other methods. In the ``tree standing alongside road'' example, RLFSeg successfully filters out the empty spaces between the tree branches. While ClipSeg can also filter these hollow regions to some extent, it introduces other artifacts. In contrast, LD-ZNet recognizes the overall tree structure well but fails to handle the hollow areas within the branches.

\begin{figure}[t]
    \centering
    \includegraphics[width=0.9\linewidth]{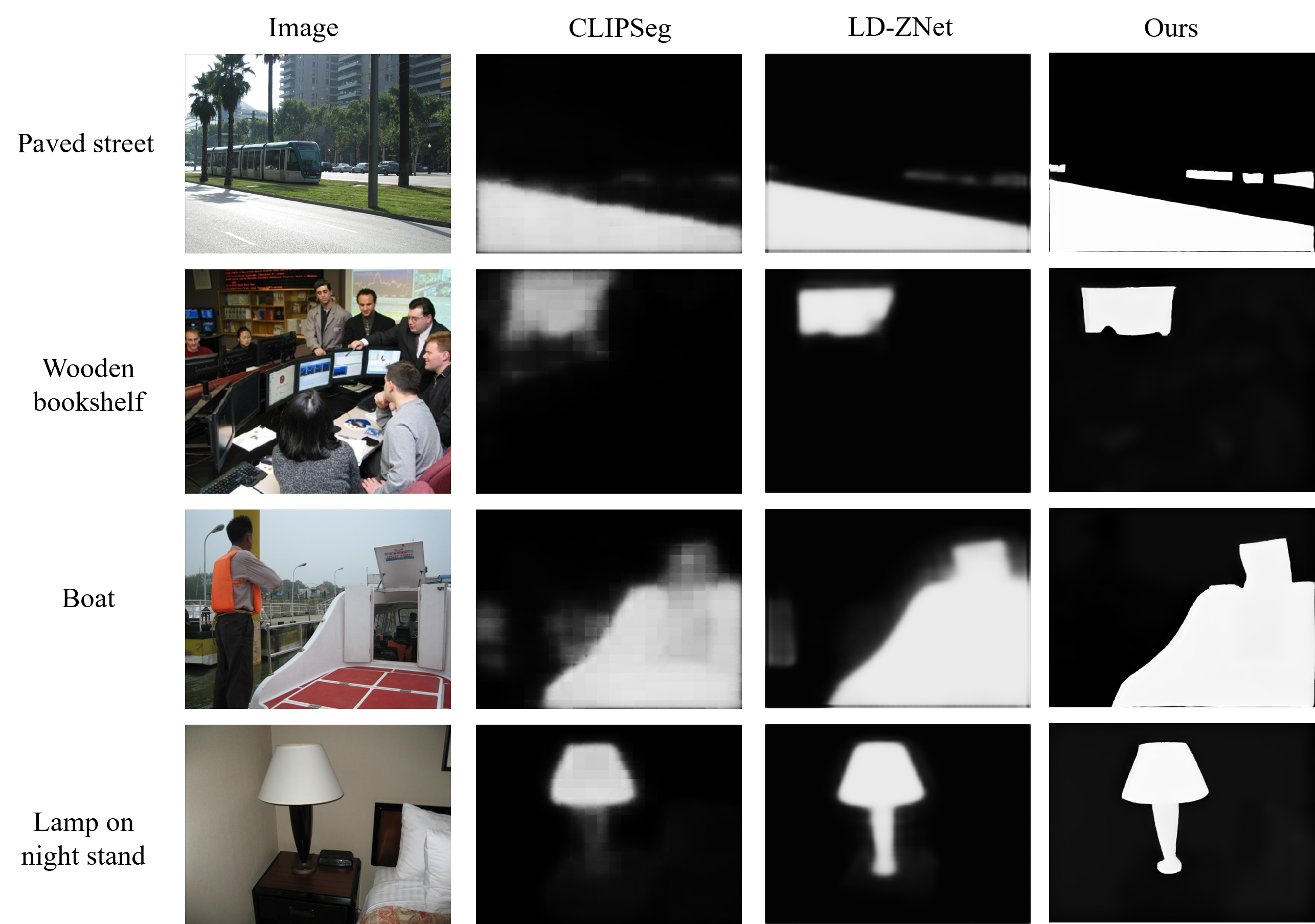}
    \caption{\textbf{Qualitative comparison of mask boundaries.} Leveraging the advantages of applying Rectified Flow to image segmentation tasks, our proposed method generates significantly sharper and better-defined edges compared to competing approaches.}
    \label{fig:boundaries}
\end{figure}

\textbf{Well-defined boundaries.} As can be seen in Figure \ref{fig:boundaries} Unlike most conventional segmentation methods, the masks generated by RLFSeg exhibit remarkably sharp and well-defined boundaries, a characteristic reminiscent of generative models' proficiency in synthesizing high-frequency details. We attribute this advantage to our training strategy of learning the ground truth distribution in the latent space, which allows our model to distinguish itself from numerous other segmentation approaches.

% \begin{figure*}[t]
%     \centering
%     \begin{minipage}[t]{0.48\linewidth}
%         \centering
%         \includegraphics[width=\linewidth]{boundaries1.png}
%         \caption{\textbf{Qualitative comparison of mask boundaries.} Our proposed method generates significantly sharper and better-defined edges compared to competing approaches.}
%         \label{fig:boundaries}
%     \end{minipage}
%     \hfill
%     \begin{minipage}[t]{0.48\linewidth}
%         \centering
%         \includegraphics[width=\linewidth]{step_visual1.png}
%         \caption{\textbf{Visualization of the path-crossing issues.} This figure shows that at \(t=0.8\), the predicted direction of v changes drastically compared to its direction at \(t=1.0\)}
%         \label{fig:path-crossing}
%     \end{minipage}
%     \vspace{-2.0em}
% \end{figure*}

\begin{figure}[t]
    \centering
    \includegraphics[width=0.9\linewidth]{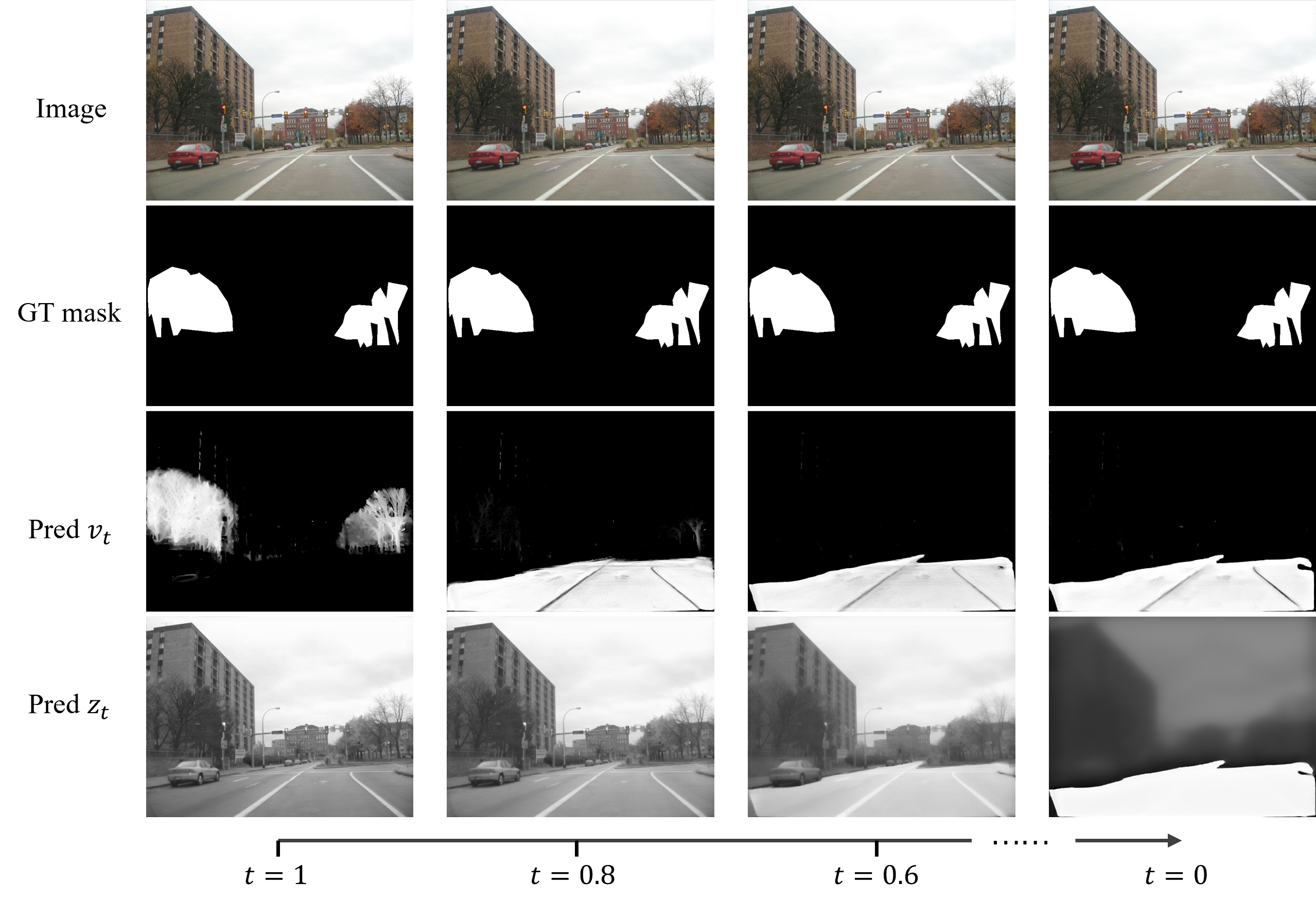}
    \caption{\textbf{Visualization of the path-crossing issues.} This figure shows that at \(t=0.8\), the predicted direction of \(v\) changes drastically compared to its direction at \(t=1.0\).}
    \label{fig:path-crossing}
    \vskip -1.5em
\end{figure}

\textbf{Path-crossing issues\label{path_crossing}.} When utilizing Rectified Flow, the significant distributional overlap between the source $z_0$ (RGB color images) and the target $z_1$ (grayscale masks) can lead to path-crossing issues which is a phenomenon where a single initial state's generative trajectory bifurcates, leading to multiple distinct terminal points, despite the deterministic nature of the semantic segmentation task where a unique mapping from image to mask should exist. We found this problem to be particularly acute in the early stages of the trajectory, specifically between timesteps $t=0.9$ and $t=0.7$. During this critical phase, the model is determining its initial direction, making it highly susceptible to interference from the overlapping distributions. This can cause incorrect predictions, such as confusing the white foreground of the mask with bright white regions in the original image. Consequently, increasing the number of sampling steps can sometimes degrade performance. Figure \ref{fig:path-crossing}. presents a particularly extreme case of this failure mode; however, it is important to note that most cases are not this severe.

\section{CONCLUSION}
In our work, we have proposed RLFSeg, a method that integrates the traditional task of semantic segmentation with flow matching. Within this framework, the segmentation task is seamlessly fused with the Latent Diffusion Model (LDM) architecture, in contrast to previous works that often rigidly employed diffusion models as internal feature extractors. This approach allows us to bypass the problem of timestep selection for the image segmentation task and avoids the dependency on random sampling noise inherent to original diffusion models, leading to a more streamlined and elegant process design. Through comprehensive experiments, we have demonstrated that our method not only achieves strong results on in-distribution test sets but also exhibits superior generalization capabilities compared to prior works in this research area.

% \textbf{Reproducibility Statement.} To ensure the full reproducibility of our findings and to facilitate future research in this area, we have made comprehensive efforts to document our work. Section~\ref{experimental_details} detail the availability of our data, experimental setup, and all necessary parameters.

%%
%% The acknowledgments section is defined using the "acks" environment
%% (and NOT an unnumbered section). This ensures the proper
%% identification of the section in the article metadata, and the
%% consistent spelling of the heading.
% \begin{acks}
% To Robert, for the bagels and explaining CMYK and color spaces.
% \end{acks}

%%
%% The next two lines define the bibliography style to be used, and
%% the bibliography file.
\bibliographystyle{ACM-Reference-Format}
\bibliography{icmr}

% \section{Research Methods}

% \subsection{Part One}

% Lorem ipsum dolor sit amet, consectetur adipiscing elit. Morbi
% malesuada, quam in pulvinar varius, metus nunc fermentum urna, id
% sollicitudin purus odio sit amet enim. Aliquam ullamcorper eu ipsum
% vel mollis. Curabitur quis dictum nisl. Phasellus vel semper risus, et
% lacinia dolor. Integer ultricies commodo sem nec semper.

% \subsection{Part Two}

% Etiam commodo feugiat nisl pulvinar pellentesque. Etiam auctor sodales
% ligula, non varius nibh pulvinar semper. Suspendisse nec lectus non
% ipsum convallis congue hendrerit vitae sapien. Donec at laoreet
% eros. Vivamus non purus placerat, scelerisque diam eu, cursus
% ante. Etiam aliquam tortor auctor efficitur mattis.

% \section{Online Resources}

% Nam id fermentum dui. Suspendisse sagittis tortor a nulla mollis, in
% pulvinar ex pretium. Sed interdum orci quis metus euismod, et sagittis
% enim maximus. Vestibulum gravida massa ut felis suscipit
% congue. Quisque mattis elit a risus ultrices commodo venenatis eget
% dui. Etiam sagittis eleifend elementum.

% Nam interdum magna at lectus dignissim, ac dignissim lorem
% rhoncus. Maecenas eu arcu ac neque placerat aliquam. Nunc pulvinar
% massa et mattis lacinia.

\end{document}